\documentclass{llncs}
\usepackage{url}
\usepackage{paralist}
\usepackage{amsmath}
\usepackage{booktabs}
\usepackage{multirow}
\usepackage{lscape}
\usepackage[final]{graphicx}
\usepackage{array}
\usepackage[T1]{fontenc}
\usepackage{hyperref}
\usepackage{subfigure}
\usepackage[all]{hypcap}
\usepackage[sort]{cite}

\newcommand{\cnf}{\textsc{cnf}}
\newcommand{\sat}{\textsc{sat}}
\newcommand{\csp}{\textsc{csp}}
\newcommand{\xcsp}{\textsc{xcsp}}

\newcommand{\cputime}{\textsc{CPU-Time}}
\newcommand{\true}{\emph{true}}
\newcommand{\false}{\emph{false}}
\newcommand{\satzilla}{\textsc{SATzilla}}
\newcommand{\cphydra}{\textsc{CPhydra}}

\newcommand{\isac}{\textsc{isac}}
\newcommand{\minisat}{\texttt{MiniSat}}

\newcommand{\mistral}{\texttt{Mistral}}
\newcommand{\clasp}{\texttt{Clasp}}
\newcommand{\sugar}{\texttt{Sugar}}
\newcommand{\azucar}{\texttt{Azucar}}

\newcommand{\abscon}{\texttt{Abscon}}
\newcommand{\choco}{\texttt{Choco}}

\newcommand{\satforj}{\texttt{SAT4J}}

\newcommand{\gecode}{\texttt{Gecode}}

\newcommand{\myparagraph}[1]{\medskip\noindent\textbf{#1.}}

\hypersetup{colorlinks,
	    citecolor=black,
	    filecolor=black,
	    linkcolor=black,
	    urlcolor=black}

\begin{document}

\title{Proteus: A Hierarchical Portfolio\\ of Solvers and Transformations}
\author{Barry Hurley \and Lars Kotthoff \and Yuri Malitsky \and Barry O'Sullivan}
\institute{Insight Centre for Data Analytics\\
Department of Computer Science, University College Cork, Ireland\\
\email{\{b.hurley\textbar l.kotthoff\textbar y.malitsky\textbar b.osullivan\}@4c.ucc.ie}}
\maketitle

\begin{abstract}
In recent years, portfolio approaches to solving SAT problems and CSPs have
 become increasingly common. There are also a number of different encodings for
 representing CSPs as SAT instances.
In this paper, we leverage advances in both SAT and CSP solving to
 present a novel hierarchical portfolio-based approach to CSP solving, which
  we call Proteus, that does not rely purely on CSP solvers.
Instead, it may decide that it is best to encode a CSP problem instance
 into SAT, selecting an appropriate encoding and a corresponding SAT solver.
Our experimental evaluation used an instance of Proteus that involved
 four CSP solvers, three SAT encodings, and six SAT solvers, evaluated on 
 the most challenging problem instances from the CSP solver competitions,
 involving global and intensional constraints.
We show that significant performance improvements can be achieved by
 Proteus obtained by exploiting alternative view-points and solvers for
 combinatorial problem-solving.
\end{abstract}

\section{Introduction}

The pace of development in both \csp\ and \sat\ solver technology has been
rapid. Combined with portfolio and algorithm selection technology impressive
performance improvements over systems that have been developed only a few years
previously have been demonstrated. Constraint satisfaction problems and
satisfiability problems are both NP-complete and, therefore, there exist
polynomial-time transformations between them. We can leverage this fact to
convert \csp s into \sat\ problems and solve them using \sat\ solvers.


In this paper we exploit the fact
 that different \sat\ solvers have different performances on
different encodings of the same \csp. In fact, the particular choice of encoding
that will give good performance with a particular \sat\ solver is dependent on
the problem instance to be solved. We show that, in addition to using dedicated
\csp\ solvers, to achieve the best performance for solving a \csp\ the best
course of action might be to translate it to \sat\ and solve it using a \sat\
solver. We name our approach Proteus, after the Greek god Proteus, the
shape-shifting water deity that can foretell the future.

Our approach offers a novel perspective on using \sat\ solvers for constraint
solving.
The idea of solving \csp s as \sat\ instances is not new; the
 solvers
\sugar, \azucar, and \texttt{CSP2SAT4J} are three examples of \sat-based
 \csp\ solving.
\sugar~\cite{TamuraTB:CSC:2008} has been very competitive in recent
 \csp\ solver competitions. It converts the \csp\ to \sat\ using a specific
  encoding, known as the order encoding, which will be discussed in more detail
  later in this paper.
\azucar~\cite{TanjoTB:SAT:2012} is a related \sat-based
\csp\ solver that uses the compact order encoding. However,
 both \sugar\ and \azucar\ use a single predefined solver to solve the encoded
 \csp\ instances.
Our work does not assume that conversion using a specific encoding to \sat\ is the best way of
solving a problem, but considers multiple candidate encodings and solvers.
%
\texttt{CSP2SAT4J}~\cite{LeBerreLynce:CP08} uses the \satforj\
library as its
\sat\ back-end and a set of static rules to choose either the direct or the
support encoding for each constraint. For intensional and extensional binary
constraints that specify the supports, it uses the support encoding. For all
other constraints, it uses the direct encoding. Our approach does not have
predefined rules but instead chooses the encoding and solver  based
on features of the problem instance to solve.

Our approach employs algorithm selection techniques to dynamically choose
whether to translate to \sat, and if so, which \sat\ encoding and solver to use,
 otherwise it selects which \csp\ solver to use. There
has been a great deal of research in the area of algorithm selection and
portfolios; we refer the reader to a recent survey of this
work~\cite{DBLP:journals/corr/abs-1210-7959}. We note three contrasting example
approaches to algorithm selection for the constraint satisfaction and
satisfiability problems: \cphydra\ (\csp), \satzilla\ (\sat), and
\isac\ (\sat).
\cphydra~\cite{OMahony:2008vs} contains an algorithm portfolio of \csp\ solvers
which partitions \cputime\ between components of the portfolio in
order to maximize the probability of solving a given problem instance within a fixed
time limit. \satzilla~\cite{Xu:2008:SA}, at its core, uses
cost-sensitive decision forests that vote on the \sat\ solver to use for an
instance. In addition to that, it contains a number of practical optimizations,
for example running a pre-solver to quickly solve the easy instances.
\isac~\cite{DBLP:conf/ecai/KadiogluMST10} is a cluster-based approach that
groups instances based on their features and then finds the best solver for each
cluster.
The Proteus approach is not a straightforward application of portfolio techniques. In
particular, there is a series of decisions to make that affect not only the
solvers that will be available, but also the information that can be used to
make the decision. Because of this, the different choices of conversions,
encodings and solvers cannot simply be seen as different algorithms or different
configurations of the same algorithm.

The remainder of this paper is organised as follows.
Section~\ref{sec-motivation} motivates the need to choose the representation
and solver in combination.
In Section~\ref{sec-background} we summarise the necessary background on \csp\ and
 \sat\ to make the paper self-contained and
present an overview of the main \sat\ encodings of \csp s.
The detailed evaluation of our portfolio is presented in
Section~\ref{sec-evaluation}.
 We create a portfolio-based approach to
\csp\ solving that employs four \csp\
solvers, three \sat\ encodings, and six \sat\ solvers.
Finally, we conclude in Section~\ref{sec-conclusions}.

\section{Multiple Encodings and Solvers}
\label{sec-motivation}

To motivate our work, we performed a detailed investigation for two solvers to
assess the relationship between solver and problem encoding with
 features of the problem to be
solved. For this experiment we considered uniform random binary (\textsc{urb}) \csp s with a
fixed number of variables, domain size and number of constraints, and varied the
constraint tightness. The constraint tightness $t$ is a measure of the
proportion of forbidden to allowed possible assignments to the variables in the
scope of the constraint. We vary it from 0 to 1, where 0 means that all
assignments are allowed and 1 that no assignments are part of a solution, in
increments of 0.005. At each tightness the mean run-time of the solver on 100
random \csp\ instances is reported. Each instance contains 30 variables with
domain size 20 and 300 constraints. This allowed us to study the performance of
\sat\ encodings and solvers across the phase
transition.

\begin{figure}[t]
\centering
\subfigure[Performance using \minisat.\label{fig-minisat}]
{\includegraphics[width=0.6\textwidth]{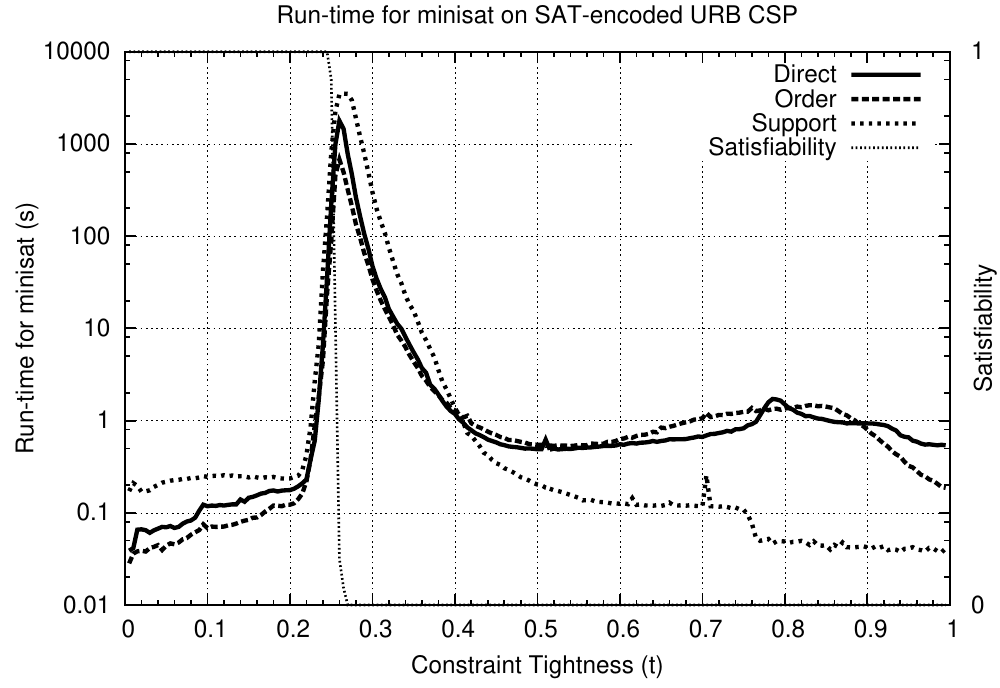}}
\subfigure[Performance using \clasp.\label{fig-clasp}]
{\includegraphics[width=0.6\textwidth]{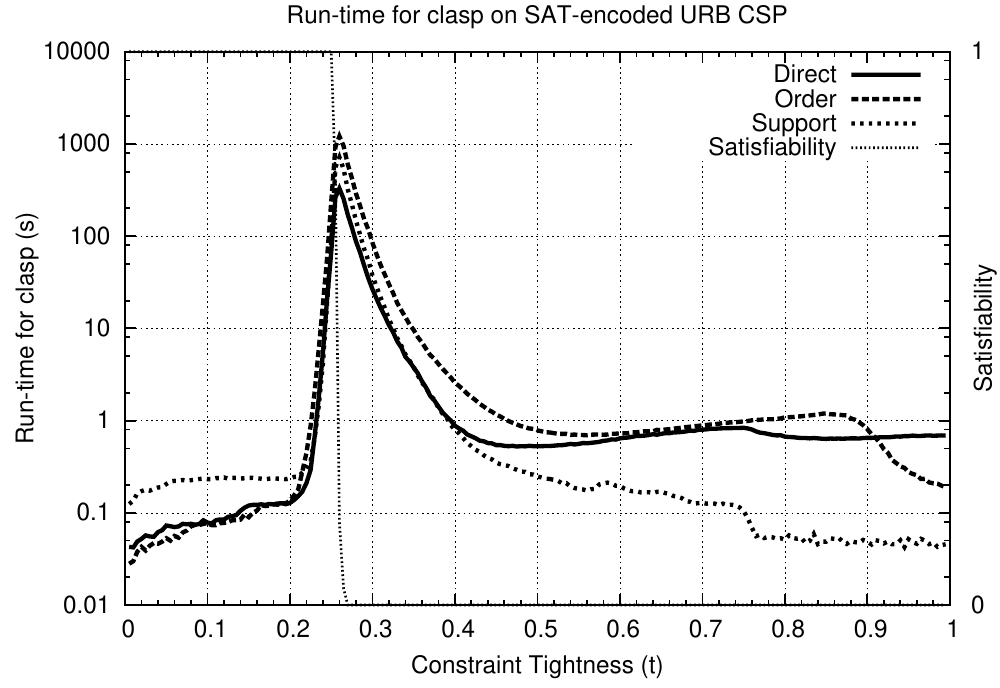}}
\caption{\minisat\ and \clasp\ on random binary \csp s.}
\label{fig:saturb}
\end{figure}

Figure~\ref{fig:saturb} plots the run-time for \minisat\ and \clasp\ on
uniformly random binary \csp s that have been translated to \sat\ using three
different encodings. Observe that in Figure~\ref{fig-minisat} there is a
distinct difference in the performance of \minisat\ on each of the encodings,
sometimes an order of magnitude. Before the phase transition, we see
that the order encoding achieves the best performance and
maintains this until the phase transition. Beginning at constraint tightness
0.41, the order encoding gradually starts achieving poorer performance and the
support encoding now achieves the best performance.

Notably, if we rank the
encodings based on their performance, the ranking changes after the phase
transition. This illustrates that there is not just a single encoding that
will perform best overall and that the choice of encoding matters, but also that
this choice is dependent on problem characteristics such as constraint
tightness.

Around the phase transition, we observe contrasting performance for \clasp, as
illustrated in Figure~\ref{fig-clasp}. Using \clasp, the ranking of encodings
around the phase transition is $\text{direct} \succ \text{support} \succ
\text{order}$; whereas for \minisat\ the ranking is $\text{order} \succ
\text{direct} \succ \text{support}$. Note also that the peaks at the
phase transition differ in magnitude between the two solvers. These differences
underline the importance of the choice of solver, in particular in conjunction
with the choice of encoding -- making the two choices in isolation does not
consider the interdependencies that affect performance in practice.

\begin{figure}[t!]
\centering
\includegraphics[width=0.7\textwidth]{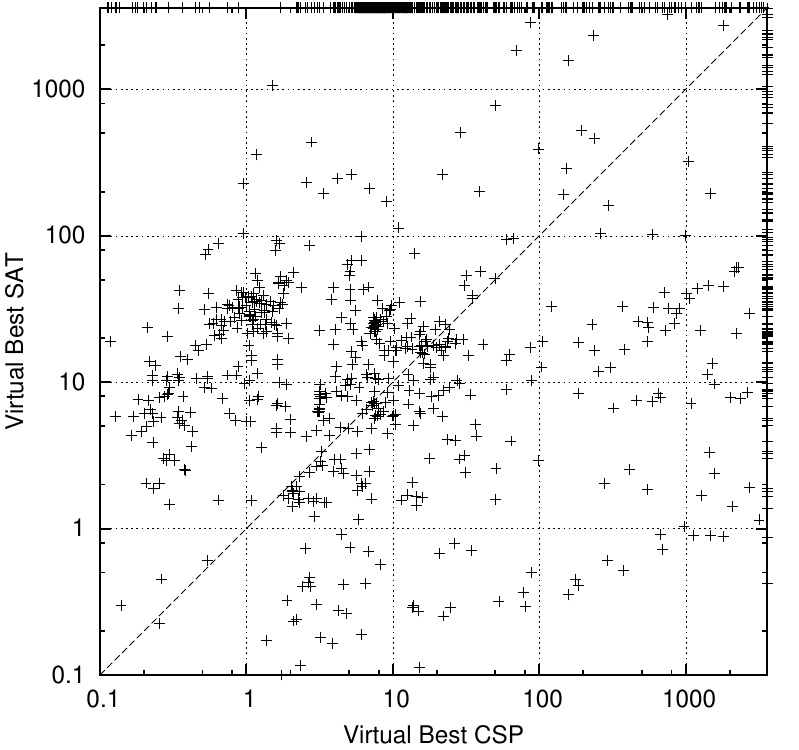}
\caption{Performance of the virtual best \csp\ portfolio and the virtual best
\sat-based portfolio. Each point represents the time in seconds of the two approaches.
A point below the dashed line indicates that the virtual best \sat\ portfolio was
quicker, whereas a point above means the virtual best \csp\ portfolio was quicker.
Clearly the two approaches are complementary: there
are numerous instances for which a \sat-based approach does not perform well or
fails to solve the instance but a \csp\ solver does extremely well,
 and vice-versa.}
\label{fig:scattercspsat}
\end{figure}

In addition to the random \csp\ instances, our analysis also comprises 1493
 challenging benchmark problem instances from the \csp\ solver competitions that
 involve global and intensional constraints. Figure~\ref{fig:scattercspsat}
illustrates the respective performance of the best \csp-based and \sat-based 
 methods on
these instances. Unsurprisingly the dedicated \csp\ methods often achieve the
best performance. There are, however, numerous cases where considering \sat-based methods
has the potential to yield significant performance improvements. In particular,
there are a number of instances that are unsolved by any \csp\ solver but can be
solved quickly using \sat-based methods. The Proteus approach aims to unify the best of
both worlds and take advantage of the potential performance gains.

\section{Background}
\label{sec-background}

\subsection{The Constraint Satisfaction Problem}

Constraint satisfaction problems (\csp) are a natural means of expressing and
reasoning about combinatorial problems.
 They
have a large number of practical applications such as scheduling, planning,
vehicle routing, configuration, network design, routing and wavelength
assignment~\cite{CPHandbook}. An instance of a \csp\ is represented by a set of
variables, each of which can be assigned a value from its domain. The
assignments to the variables must be consistent with a set of constraints, where
each constraint limits the values that can be assigned to variables.


Finding a solution to a \csp\ is typically done using systematic search based
on backtracking. Because the general problem is NP-complete, systematic search
algorithms have exponential worst-case run times, which has the effect of
limiting the scalability of these methods. However, thanks to the development of
effective heuristics and a wide variety of solvers with different strengths and
weaknesses, many problems can be solved efficiently in practice.

\subsection{The Satisfiability Problem\label{subsec:sat}}

The satisfiability problem (\sat) consists of a set of Boolean variables and a
propositional formula over these variables. The task is to decide whether or not
there exists a truth assignment to the variables such that the propositional
formula evaluates to \true, and, if this is the case, to find this assignment.

\sat\ instances are usually expressed in conjunctive normal form (\cnf). The
representation consists of a conjunction of \emph{clauses}, where each clause is
a disjunction of \emph{literals}. A literal is either a variable or its
negation. Each clause is a logical \emph{or} of its literals and the formula is
a logical \emph{and} of each clause. The following \sat\ formula is in \cnf:
\[
(x_1 \vee x_2 \vee \neg x_4) \wedge (\neg x_2 \vee \neg x_3) \wedge (x_3 \vee
x_4)
\]
This instance consists of four \sat\ variables. One assignment to the
variables which would satisfy the above formula would be to set $x_1=\true$,
$x_2=\false$, $x_3=\true$ and $x_4= \true$.

\sat, like \csp, has a variety of practical real world applications such as
hardware verification, security protocol analysis, theorem proving, scheduling,
routing, planning, digital circuit design~\cite{HandbookOfSAT2009}. The
application of \sat\ to many of these problems is made possible by
transformations from representations like the constraint satisfaction problem.
We will study three transformations into \sat\ that can benefit from this large
collection of solvers.

The following sections explain the direct, support, and direct-order encodings that we
use. We will use the following notation. The set of \csp\ variables is
represented by the set $\mathcal{X}$. We use uppercase letters to denote \csp\
variables in $\mathcal{X}$; lowercase $x_i$ and $x_v$ refer to \sat\
variables. The domain of a \csp\ variable $X$ is denoted $\text{D}(X)$ and has size $d$.

\subsection{Direct Encoding}

Translating a \csp\ variable $X$ into \sat\ using the \emph{direct
encoding}~\cite{Walsh:2000}, also known as the \emph{sparse encoding},
 creates a \sat\ variable for each value in its domain: $x_1, x_2, \ldots, x_d$.
 If $x_v$ is
\true\ in the resulting \sat\ formula, then $X=v$ in the \csp\ solution. This
means that in order to represent a solution to the \csp, exactly one of $x_1, x_2,
\ldots, x_d$ must be assigned \true. We add an \emph{at-least-one} clause to the
\sat\ formula for each \csp\ variable as follows:
\[
\forall{X \in \mathcal{X}}: (x_1 \vee x_2 \vee \ldots \vee x_d).
\]
Conversely, to ensure that only one of these can be set to \true, we add
\emph{at-most-one} clauses. For each pair of distinct values in the domain of
$X$, we add a binary clause to enforce that at most one of the two can be
assigned \true. The series of these binary clauses ensure that only one of the
\sat\ variables representing the variable will be assigned \true, i.e.
\[
\forall{v, w \in \text{D}(X)}: (\neg x_v \vee \neg x_w).
\]


Constraints between \csp\ variables are represented in the direct encoding by
enumerating the conflicting tuples. For binary constraints for example, we add
clauses as above to forbid both values being used at the same time for each
disallowed assignment. For a binary constraint between a pair of variables $X$
and $Y$, we add the conflict clause $(\neg x_v \vee \neg y_w)$ if the tuple
$\langle X=v, Y=w \rangle$ is forbidden.
For intensionally specified constraints, we
enumerate all possible tuples and encode the disallowed assignments.

\begin{example}[Direct Encoding]
\label{ex:simplecspdirect}
Consider a simple \csp\ with three variables $\mathcal{X} = \{X, Y, Z\}$, each
with domain $\langle 1, 2, 3 \rangle$. We have an all-different constraint over the variables: $\text{alldifferent}(X, Y, Z)$,
which we represent by encoding the pairwise dis-equalities.
Table~\ref{tab-ex-direct} shows the complete
direct-encoded \cnf\ formula for this \csp. The first 12 clauses encode the
domains of the variables, the remaining clauses encode the constraints between
$X$, $Y$, and $Z$. There is an implicit conjunction between these clauses.

\begin{table}[htbp]
\caption{An example of the direct encoding.}
\label{tab-ex-direct}
\centering
\begin{tabular}{c @{\hspace{10pt}} p{7.5cm}}
\toprule
\multirow{3}{*}{\textbf{Domain Clauses}} &
$(x_1 \vee x_2 \vee x_3)$ $(\neg x_1 \vee \neg x_2)$
$(\neg x_1 \vee \neg x_3)$ $(\neg x_2 \vee \neg x_3)$ \\
& $(y_1 \vee y_2 \vee y_3)$ $(\neg y_1 \vee \neg y_2)$
$(\neg y_1 \vee \neg y_3)$ $(\neg y_2 \vee \neg y_3)$ \\
& $(z_1 \vee z_2 \vee z_3)$ $(\neg z_1 \vee \neg z_2)$
$(\neg z_1 \vee \neg z_3)$ $(\neg z_2 \vee \neg z_3)$ \\
\midrule

\textbf{$X \neq Y$} &
$(\neg x_1 \vee \neg y_1)$ $(\neg x_2 \vee \neg y_2)$ $(\neg x_3 \vee \neg y_3)$
\\
\textbf{$X \neq Z$} &
$(\neg x_1 \vee \neg z_1)$ $(\neg x_2 \vee \neg z_2)$ $(\neg x_3 \vee \neg z_3)$
\\
\textbf{$Y \neq Z$} &
$(\neg y_1 \vee \neg z_1)$ $(\neg y_2 \vee \neg z_2)$ $(\neg y_3 \vee \neg z_3)$
\\ \bottomrule
\end{tabular}
\end{table}
\end{example}

\subsection{Support Encoding}

The \emph{support encoding}~\cite{Kasif:1990,DBLP:conf/ecai/Gent02}
 uses the same mechanism as the
direct encoding to encode \csp\ domains into \sat\ --
each value in the domain of a \csp\ variable is encoded as a \sat\
variable which represents whether or not it takes that value. However, the
support encoding differs on how the constraints between variables are encoded.
Given a constraint between two variables $X$ and $Y$, for each value $v$ in the
domain of $X$, let $S_{Y,X=v} \subset D(Y)$ be the subset of the values in the
domain of $Y$ which are consistent with assigning $X=v$. Either $x_v$ is \false\
or one of the consistent assignments from $y_1 \ldots y_d$  must be true.
This is encoded in the support clause
\[ \neg x_v \vee \left( \bigvee_{i \in S_{Y,X=v}} y_i \right). \]
Conversely, for each value $w$ in the domain of $Y$, a support clause is added for
the supported values in $X$ which are consistent with assigning $Y=w$.

\smallskip
An interesting property of the support encoding is that if a constraint has no
consistent values in the corresponding variable, a unit-clause will be
added, thereby pruning the values from the domain of a variable which cannot
exist in any solution. Also, a solution to a \sat\ formula without the
\emph{at-most-one} constraint in the support encoding represents an
arc-consistent assignment to the \csp. Unit propagation on this \sat\ instance
establishes arc-consistency in optimal worst-case time for establishing
arc-consistency~\cite{DBLP:conf/ecai/Gent02}.

\begin{example}[Support Encoding]
\label{ex:simplecspsupport}
Table~\ref{tab-ex-support} gives the complete support-encoded \cnf\ formula for the
simple \csp\ given in Example~\ref{ex:simplecspdirect}. The first 12 clauses
encode the domains and the remaining ones the support clauses for the
constraints. There is an implicit conjunction between clauses.

\begin{table}[htbp]
\caption{An example of the support encoding.}
\label{tab-ex-support}
\centering
\begin{tabular}{c @{\hspace{10pt}} p{7.5cm}}
\toprule
\multirow{3}{*}{\textbf{Domain Clauses}} &
$(x_1 \vee x_2 \vee x_3)$ $(\neg x_1 \vee \neg x_2)$
$(\neg x_1 \vee \neg x_3)$ $(\neg x_2 \vee \neg x_3)$ \\
& $(y_1 \vee y_2 \vee y_3)$ $(\neg y_1 \vee \neg y_2)$
$(\neg y_1 \vee \neg y_3)$ $(\neg y_2 \vee \neg y_3)$ \\
& $(z_1 \vee z_2 \vee z_3)$ $(\neg z_1 \vee \neg z_2)$
$(\neg z_1 \vee \neg z_3)$ $(\neg z_2 \vee \neg z_3)$ \\
\midrule

\multirow{2}{*}{\textbf{$X \neq Y$}}
& $(\neg x_1 \vee y_2 \vee y_3)$ $(\neg x_2 \vee y_1 \vee y_3)$
$(\neg x_3 \vee y_1 \vee y_2)$ \\
& $(\neg y_1 \vee x_2 \vee x_3)$ $(\neg y_2 \vee x_1 \vee x_3)$
$(\neg y_3 \vee x_1 \vee x_2)$ \\
\midrule

\multirow{2}{*}{\textbf{$X \neq Z$}}
& $(\neg x_1 \vee z_2 \vee z_3)$ $(\neg x_2 \vee z_1 \vee z_3)$
$(\neg x_3 \vee z_1 \vee z_2)$ \\
& $(\neg z_1 \vee x_2 \vee x_3)$ $(\neg z_2 \vee x_1 \vee x_3)$
$(\neg z_3 \vee x_1 \vee x_2)$ \\
\midrule

\multirow{2}{*}{\textbf{$Y \neq Z$}}
& $(\neg y_1 \vee z_2 \vee z_3)$ $(\neg y_2 \vee z_1 \vee z_3)$
$(\neg y_3 \vee z_1 \vee z_2)$ \\
& $(\neg z_1 \vee y_2 \vee y_3)$ $(\neg z_2 \vee y_1 \vee y_3)$
$(\neg z_3 \vee y_1 \vee y_2)$ \\
\bottomrule
\end{tabular}
\end{table}
\end{example}

\subsection{Order Encoding}
Unlike the direct and support encoding, which model $X=v$ as a \sat\ variable
for each value $v$ in the domain of $X$, the order encoding (also known as the
regular encoding~\cite{DBLP:conf/sat/AnsoteguiM04}) creates \sat\
variables to represent $X \leq v$. If $X$ is less than or equal to $v$ (denoted
$x_{\leq v}$), then $X$ must also be less than or equal to $v+1$ ($x_{\leq
v+1}$). Therefore, we add clauses to enforce this consistency across the domain
as follows:
\[
\forall_{v}^{d-1} : (\neg x_{\leq v} \vee x_{\leq v+1}).
\]
This linear number of clauses 
is all that is needed to encode the domain of a \csp\ variable into \sat\ in
the order encoding. In contrast, the direct and support encodings require a
quadratic number of clauses in the domain size.


The order encoding is naturally suited to modelling inequality constraints. To
state $X \leq 3$, we would just post the unit clause $(x_{\leq 3})$. If we want to
model the constraint $X=v$, we could rewrite it as $(X \leq v \wedge X \geq v)$.
$X \geq v$ can then be rewritten as $\neg X \leq (v-1)$. To state that $X=v$ in
the order encoding, we would encode $(x_{\leq v} \wedge \neg x_{\leq v-1})$. A
conflicting tuple between two variables, for example $\langle X=v, Y=w \rangle$
can be written in propositional logic and simplified to a \cnf\ clause
using De Morgan's Law:
\begin{align*}
\neg ((x_{\leq v} \wedge x_{\geq v}) &\wedge
      (y_{\leq w} \wedge y_{\geq w})) \\
\neg ((x_{\leq v} \wedge \neg x_{\leq v-1}) &\wedge 
      (y_{\leq w} \wedge \neg y_{\leq w-1}) ) \\
\neg (x_{\leq v} \wedge \neg x_{\leq v-1}) &\vee
\neg (y_{\leq w} \wedge \neg y_{\leq w-1}) \\
(\neg x_{\leq v} \vee x_{\leq v-1} &\vee \neg y_{\leq w} \vee y_{\leq w-1})
\end{align*}

\begin{example}[Order Encoding]
\label{ex:simplecsporder}
Table~\ref{tab-ex-order} gives the complete order-encoded \cnf\ formula for the
simple \csp\ specified in Example~\ref{ex:simplecspdirect}. There is an implicit
conjunction between clauses in the notation.

\begin{table}[htbp]
\caption{An example of the order encoding.}
\label{tab-ex-order}
\centering
\begin{tabular}{c @{\hspace{10pt}} p{7.5cm}}
\toprule
\multirow{3}{*}{\textbf{Domain Clauses}}
& $(\neg x_{\leq 1} \vee x_{\leq 2})$ $(\neg x_{\leq 2} \vee x_{\leq 3})$
$(x_{\leq 3})$ \\
& $(\neg y_{\leq 1} \vee y_{\leq 2})$ $(\neg y_{\leq 2} \vee y_{\leq 3})$
$(y_{\leq 3})$ \\
& $(\neg z_{\leq 1} \vee z_{\leq 2})$ $(\neg z_{\leq 2} \vee z_{\leq 3})$
$(z_{\leq 3})$ \\
\midrule

\multirow{3}{*}{\textbf{$X \neq Y$}}
& $(\neg x_{\leq 1} \vee \neg y_{\leq 1})$ \\
& $(\neg x_{\leq 2} \vee x_{\leq 1} \vee \neg y_{\leq 2} \vee y_{\leq 1})$ \\
& $(\neg x_{\leq 3} \vee x_{\leq 2} \vee \neg y_{\leq 3} \vee y_{\leq 2})$ \\
\midrule

\multirow{3}{*}{\textbf{$X \neq Z$}}
& $(\neg x_{\leq 1} \vee \neg z_{\leq 1})$ \\
& $(\neg x_{\leq 2} \vee x_{\leq 1} \vee \neg z_{\leq 2} \vee z_{\leq 1})$ \\
& $(\neg x_{\leq 3} \vee x_{\leq 2} \vee \neg z_{\leq 3} \vee z_{\leq 2})$ \\
\midrule

\multirow{3}{*}{\textbf{$Y \neq Z$}}
& $(\neg y_{\leq 1} \vee \neg z_{\leq 1})$ \\
& $(\neg y_{\leq 2} \vee y_{\leq 1} \vee \neg z_{\leq 2} \vee z_{\leq 1})$ \\
& $(\neg y_{\leq 3} \vee y_{\leq 2} \vee \neg z_{\leq 3} \vee z_{\leq 2})$ \\
\bottomrule
\end{tabular}
\end{table}
\end{example}

\subsection{Combining the Direct and Order Encodings}

The direct encoding and the order encoding can be combined to produce a
potentially more compact encoding. A variable's domain is encoded in both
representations and clauses are added to chain between them. This gives
flexibility in the representation of each constraint. Here, we choose the
encoding which gives the most compact formula. For example, for inequalities we
use the order encoding since it is naturally suited, but for a (dis)equality we
would use the direct encoding. This encoding is referred to as direct-order
throughout the paper.

\subsection{Algorithm Portfolios}

The Algorithm Selection Problem \cite{rice_algorithm_1976} is to select the most
appropriate algorithm for solving a particular problem. It is especially
relevant in the context of algorithm portfolios
\cite{huberman_economics_1997,gomes_algorithm_2001}, where a single solver is
replaced with a set of solvers and a mechanism for selecting a subset to use on
a particular problem.

Algorithm portfolios have been used with great success for solving both \sat\ and
\csp\ instances in systems such as \satzilla~\cite{Xu:2008:SA},
\isac~\cite{DBLP:conf/ecai/KadiogluMST10} or \cphydra~\cite{OMahony:2008vs}. Most
approaches are similar in that they relate the characteristics of a problem to
solve to the performance of the algorithms in the portfolio. The aim of an
algorithm selection model is to provide a prediction as to which algorithm
should be used to solve the problem. The model is usually induced using some
form of machine learning.

There are three main approaches to using machine learning to build algorithm
selection models. First, the problem of predicting the best algorithm can be
treated as a classification problem where the label to predict is the algorithm.
Second, the training data can be clustered and the algorithm with the best
performance on a particular cluster assigned to it. The cluster membership of
any new data decides the algorithm to use. Finally, regression models can be
trained to predict the performance of each portfolio algorithm in isolation. The
best algorithm for a problem is chosen based on the predicted performances.

Our approach makes a series of decisions -- whether a problem should
be solved as a \csp\ or a \sat\ problem, which encoding should be used for
converting into \sat, and finally which solver should be assigned to tackle the
problem. Approaches that make a series of decisions are usually referred to as
hierarchical models. \cite{xu_hierarchical_2007} and \cite{haim_restart_2009}
use hierarchical models in the context of a \sat\ portfolio. They first predict
whether the problem to be solved is expected to be satisfiable or not and then
choose a solver depending on that decision. Our approach is closer to
\cite{gent_machine_2010}, which first predicts what level of consistency the
\texttt{alldifferent} constraint should achieve before deciding on its
implementation.

To the best of our knowledge, no portfolio approach that potentially transforms
the representation of a problem in order to be able to solve it more efficiently
exists at present.

\section{Experimental Evaluation}
\label{sec-evaluation}

\subsection{Setup}

The hierarchical model we present in this paper consists of a number of layers
to determine how the instance should be solved. At the top level, we decide
whether to solve the instance using as a \csp\ or using a
\sat-based method. If we choose to leave the problem as a \csp, then one of the
dedicated \csp\ solvers must be chosen. Otherwise, we must choose
the \sat\ encoding to apply, followed by the choice of \sat\ solver to run on
the \sat-encoded instance.

Each decision of the hierarchical approach aims to choose the direction which
has the potential to achieve the best performance in that sub-tree. For example,
for the decision to choose whether to solve the instance using a \sat-based
method or not, we choose the \sat-based direction if there is a \sat\
solver and encoding that will perform faster than any \csp\
solver would. Whether this particular encoding-solver combination will be
selected subsequently depends on the performance of the algorithm selection
models used in that sub-tree of our decision mechanism. For regression models,
the training data is the best performance of any solver under that branch of the
tree. For classification models, it is the label of the sub-branch with the
virtual best performance.

This hierarchical approach presents the opportunity to employ different decision
mechanisms at each level. We consider 6 regression, 19 classification, and 3
clustering algorithms, which are listed below. For each of these algorithms, we
evaluate the performance using 10-fold cross-validation. The dataset is split
into 10 partitions with approximately the same size and the same distribution of
the best solvers. One partition is used for testing and the remaining 9
partitions as the training data for the model. This process is repeated with a
different partition considered for testing each time until every partition has
been used for testing. We measure the performance in terms of PAR10 score.
The PAR10 score for an instance is the time it takes the solver to solve the
instance, unless the solver times out. In this case, the PAR10 score is ten
times the timeout value. The sum over all instances is divided by the number of
instances.

\myparagraph{Instances} In our evaluation, we consider \csp\ problem
instances from the \csp\ solver competitions~\cite{cspcompbenchmarks}.
Of these, we consider all instances defined using global and intensional constraints
 that are not trivially
solved during 2 seconds of feature computation. We also exclude all instances
which were not solved by any \csp\ or \sat\ solver within the time limit of 1
hour. Altogether, we obtain 1,493
non-trivial instances from problem classes such as Timetabling, Frequency
Assignment, Job-Shop, Open-Shop, Quasi-group, Costas Array, Golomb Ruler, Latin
Square, All Interval Series, Balanced Incomplete Block Design, and many others.
This set includes both small and large arity constraints and all of the global
constraints used during the \csp\ solver competitions: all-different, element,
weighted sum, and cumulative.


For the \sat-based approaches, Numberjack~\cite{DBLP:conf/cpaior/HebrardOO10}
was used to translate a \csp\ instance specified in \xcsp\
format~\cite{DBLP:journals/corr/abs-0902-2362} into \sat\ (\cnf).


\myparagraph{Features}
A fundamental requirement of any machine learning algorithm is a set of
representative features. We explore a number of different feature sets to train
our models:
\begin{inparaenum}[\itshape i\upshape)]
\item features of the original \csp\ instance, 
\item features of the direct-encoded \sat\ instance,
\item features of the support-encoded \sat\ instance,
\item features of the direct-order-encoded \sat\ instance and
\item a combination of all four feature sets.
\end{inparaenum}
These features are described in further detail below.

We computed the 36 features used in \cphydra\ for each \csp\ instance using
\mistral; for reasons of space we will not enumerate them all here.
The set includes static features like statistics about the types of constraints
used, average and maximum domain size; and dynamic statistics recorded by
running \mistral\ for 2 seconds: average and standard deviation of variable
weights, number of nodes, number of propagations and a few others. Instances
which are solved by \mistral\ during feature computation are filtered out from
the dataset.


In addition to the \csp\ features, we computed the 54 \sat\ features used by
 \satzilla~\cite{Xu:2008:SA} for each of the
encoded instances and different encodings.
 The features encode a wide range of
different information on the problem such as problem size, features of the
graph-based representation, balance features, the proximity to a Horn formula,
DPLL probing features and local search probing features.

\myparagraph{Constraint Solvers}
\label{subsec-cspsolvers}
Our \csp\ models are able to choose from 4 complete \csp\ solvers:
\vspace{-5pt}
\begin{multicols}{2}
\begin{compactitem}
\item \abscon~\cite{abscon},
\item \choco~\cite{choco},
\item \gecode~\cite{gecode}, and
\item \mistral~\cite{mistral}.
\end{compactitem}
\end{multicols}

\myparagraph{Satisfiability Solvers}
\label{subsec-satsolvers}
We considered the following 6 complete \sat\ solvers:
\vspace{-5pt}
\begin{multicols}{2}
\begin{compactitem}
\item \texttt{clasp}~\cite{Gebser07clasp:a},
\item \texttt{cryptominisat}~\cite{crypto290},
\item \texttt{glucose}~\cite{audemard2013glucose},
\item \texttt{lingeling}~\cite{biere2013lingeling},
\item\texttt{riss}~\cite{manthey2013sat}, and
\item \minisat~\cite{MiniSAT}.
\end{compactitem}
\end{multicols}

\myparagraph{Learning Algorithms}
We evaluate a number of regression, classification, and clustering algorithms
using WEKA~\cite{hall_weka_2009}. All algorithms, unless otherwise stated
use the default parameters. The regression algorithms we used were LinearRegression, PaceRegression, REPTree, M5Rules, M5P, and
SMOreg. The classification algorithms were BayesNet, BFTree, ConjunctiveRule, DecisionTable, FT, HyperPipes, IBk
(nearest neighbour) with 1, 3, 5 and 10 neighbours, J48, J48graft, JRip,
LADTree, MultilayerPerceptron, OneR, PART, RandomForest, RandomForest with 99
random trees, RandomTree, REPTree, and SimpleLogistic. For clustering, we considered EM,
FarthestFirst, and SimplekMeans. The FarthestFirst and SimplekMeans algorithms
require the number of clusters to be given as input. We evaluated with multiples
of 1 through 5 of the number of solvers in the respective data set given as the
number of clusters. The number of clusters is represented by $1n$, $2n$ and so
on in the name of the algorithm, where $n$ stands for the number of solvers.

We use the LLAMA toolkit~\cite{kotthoff_llama_2013} to train and test the
algorithm selection models.

\subsection{Portfolio and Solver Results}

\begin{table*}[t]
\caption{
Performance of the learning algorithms for the hierarchical approach.
The `Category Bests' consists of the hierarchy of algorithms where at each node
of the tree of decisions we take the algorithm that achieves the best PAR10
score for that particular decision.}
\label{tab:hier}

\centering
\begin{tabular}{lrr} \toprule
Classifier & Mean PAR10 & Number Solved \\
\midrule
VBS & 97 & 1493 \\
Proteus & 1774 & 1424 \\
M5P with csp features & 2874 & 1413 \\
Category Bests & 2996 & 1411 \\
M5Rules with csp features & 3225 & 1398 \\
M5P with all features & 3405 & 1397 \\
LinearRegression with all features & 3553 & 1391 \\
LinearRegression with csp features & 3588 & 1383 \\
MultilayerPerceptron with csp features & 3594 & 1382 \\
lm with csp features & 3654 & 1380 \\
RandomForest99 with csp features & 3664 & 1379 \\
IBk10 with csp features & 3720 & 1377 \\
RandomForest99 with all features & 3735 & 1383 \\
\bottomrule
\end{tabular}

\end{table*}

\begin{figure*}[t]
\centering
\includegraphics[angle=-90,width=\textwidth]{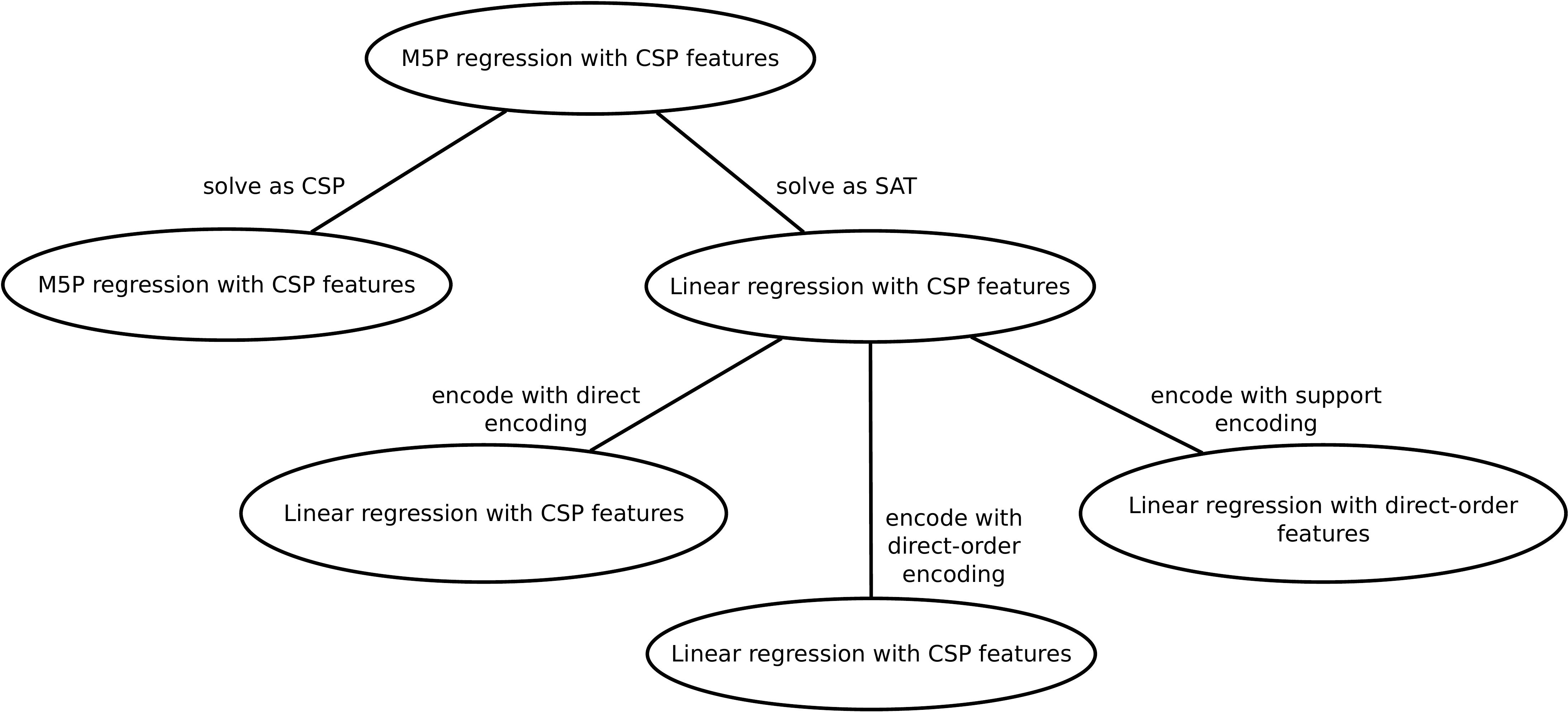}
\caption{Overview of the machine learning models used in the hierarchical
approach.\label{fig:hiermodels}}
\end{figure*}

\begin{table*}[t]
\caption{
Ranking of each classification, regression, and clustering algorithm to choose
the solving mechanism in a flattened setting. The portfolio consists of all
possible combination of the 3 encodings and the 6 \sat\ solvers and the 4
\csp\ solvers for a total of 22 solvers.}
\label{tab:hierflat}

\centering
\begin{tabular}{lrr} \toprule
Classifier & Mean PAR10 & Number Solved \\
\midrule
VBS & 97 & 1493 \\
Proteus & 1774 & 1424 \\
LinearRegression with all features & 2144 & 1416 \\
M5P with csp features & 2315 & 1401 \\
LinearRegression with csp features & 2334 & 1401 \\
lm with all features & 2362 & 1407 \\
lm with csp features & 2401 & 1398 \\
M5P with all features & 2425 & 1404 \\
RandomForest99 with all features & 2504 & 1401 \\
SMOreg with all features & 2749 & 1391 \\
RandomForest with all features & 2859 & 1386 \\
IBk3 with csp features & 2877 & 1378 \\
\bottomrule
\end{tabular}

\end{table*}

The performance of each of the 6 \sat\ solvers was evaluated on the three \sat\
encodings of 1,493 \csp\ competition benchmarks with a time-out of 1 hour and limited to 2GB of RAM.
The 4 \csp\ solvers were evaluated on the original \csp s.
Our results report the PAR10 score and number of instances solved for each of
the algorithms we evaluate.
The PAR10 is the sum of the runtimes over all instances, counting 10 times the
 timeout if that was reached.
Data was collected on a cluster of Intel Xeon
E5430 Processors (2.66Ghz) running CentOS 6.4.
This data is available online.\footnote{\url{http://4c.ucc.ie/~bhurley/proteus/}}

\label{sec-proteusinstantiation}
The performance of a number of hierarchical approaches is given in
Table~\ref{tab:hier}. The hierarchy of algorithms which produced the best
overall results for our dataset involves M5P regression with \csp\ features
at the root node to choose \sat\ or \csp, M5P regression with \csp\ features
to select the \csp\ solver, LinearRegression with \csp\ features
to select the \sat\ encoding, LinearRegression with \csp\ features 
to select the \sat\ solver for the direct encoded instance, LinearRegression
with \csp\ features to select the \sat\ solver for the direct-order encoded instance,
and LinearRegression with the direct-order features to select the \sat\
solver for the support encoded instance.
The hierarchical tree of specific machine learning approaches we found to
deliver the best overall performance on our data set is labelled Proteus and is depicted in
Figure~\ref{fig:hiermodels}.

We would like to point out that in many solver competitions the difference
between the top few solvers is fewer than 10 additional instances solved. In the
2012 \sat\ Challenge for example, the difference between the first and second
place single solver was only 3 instances and the difference among the top 4
solvers was only 8 instances. The results we present in Table~\ref{tab:hier}
are therefore very significant in terms of the gains we are able to achieve.

Our results demonstrate the power of Proteus. The performance it delivers is
very close to the virtual best (VBS), that is the best performance possible
if an oracle could identify the best choice of representation, encoding, and solver,
 on an instance by instance basis. The improvements we achieve over other
approaches are similarly impressive. The results conclusively demonstrate that
having the option to convert a \csp\ to \sat\ does not only have the potential
to achieve significant performance improvements, but also does so in practice.


An interesting observation is that the \csp\ features are consistently used in
each of the top performing approaches. One reason for this is that it is quicker
to compute only the \csp\ features instead of the \csp\ features, then
converting to \sat\ and computing the \sat\
features in addition. The additional overhead of computing \sat\ features is
worthwhile in some cases though, for example for LinearRegression, which is at
its best performance using all the different feature sets.
Note that for the best tree of models (cf.\ Figure~\ref{fig:hiermodels}), it is better
to use the features of the direct-order encoding for the decision of which
solver to choose for a support-encoded \sat\ instance despite the additional
overhead.

We also compare the hierarchical approach to that of a flattened setting with a
single portfolio of all solvers and encoding solver combinations. The
flattened portfolio includes all possible combinations of the 3 encodings and
the 6 \sat\ solvers and the 4 \csp\ solvers for a total of 22 solvers.
Table~\ref{tab:hierflat} shows these results. The regression algorithm
LinearRegression with all features gives the best performance using this
approach. However, it is significantly worse than the performance achieved by the hierarchical approach of
Proteus.

\vspace{-6pt}
\subsection{Greater than the Sum of its Parts}

Given the performance of Proteus, the question remains as to whether a different
portfolio approach that considers just \csp\ or just \sat\ solvers could do
better. Table~\ref{tab:vbs} summarizes the virtual best performance that such
portfolios could achieve. We use all the \csp\ and \sat\ solvers for the
respective portfolios to give us VB CSP and VB SAT, respectively. The former is
the approach that always chooses the best \csp\ solver for the current instance,
while the latter chooses the best \sat\ encoding/solver combination. VB Proteus
is the portfolio that chooses the best overall approach/encoding. We show the
actual performance of Proteus for comparison. Proteus is better than the virtual
bests for all portfolios that consider only one encoding. This result makes a
very strong point for the need to consider encoding and solver in combination.

Proteus outperforms four other VB portfolios.
Specifically, the VB \cphydra\ is the best possible performance that could be
 obtained from that portfolio if a perfect choice of solver was made.
Neither \satzilla\ nor \isac-based portfolios consider different \sat\ encodings.
Therefore, the best possible
performance either of them could achieve for a specific encoding is represented
 in the last three lines of Table~\ref{tab:vbs}.

These results do not only demonstrate the benefit of considering the different
ways of solving \csp s, but also eliminate the need to compare with existing
portfolio systems since we are computing the best possible performance that any
of those systems could theoretically achieve. Proteus impressively demonstrates
its strengths by significantly outperforming oracle approaches that use only a single
encoding.

\begin{table}[t]
\caption{Virtual best performances ranked by PAR10 score.}
\label{tab:vbs}
\centering
\begin{tabular}{lr@{\hspace{5pt}}r}
\toprule
Method & Mean PAR10 & Number Solved \\
\midrule
VB Proteus              &    97   & 1493 \\
Proteus                 &  1774   & 1424 \\
VB CSP                  &  3577   & 1349 \\
VB CPHydra              &  4581   & 1310 \\
VB SAT                  & 17373   &  775 \\
VB DirectOrder Encoding & 17637   &  764 \\
VB Direct Encoding      & 21736   &  593 \\
VB Support Encoding     & 21986   &  583 \\
\bottomrule
\end{tabular}
\end{table}

\vspace{-10pt}
\section{Conclusions}
\label{sec-conclusions}

We have presented a portfolio approach that does not rely on a single problem
representation or set of solvers, but leverages our ability to convert between
problem representations to increase the space of possible solving approaches. To
the best of our knowledge, this is the first time a portfolio approach like this
has been proposed. We have shown that, to achieve the best performance on a
constraint satisfaction problem, it may be beneficial to translate it to a
satisfiability problem. For this translation, it is important to choose both the
encoding and satisfiability solver in combination. In doing so, the contrasting
performance among solvers on different representations of the same problem can
be exploited. The overall performance can be improved significantly compared to
restricting the portfolio to a single problem representation.

We demonstrated empirically the significant performance improvements Proteus
 can achieve on a large set of diverse benchmarks using a portfolio based on a
range of different state-of-the-art solvers. We have investigated a range of
different \csp\ to \sat\ encodings and evaluated the performance of a large
number of machine learning approaches and algorithms. Finally, we have shown that
the performance of Proteus is close to the very best that is theoretically
possible for solving \csp s and significantly outperforms the theoretical best for portfolios
that consider only a single problem encoding.

In this work, we make a general decision to encode the entire problem using a
particular encoding. A natural extension would be to mix and vary the encoding
depending on attributes of the problem. An additional avenue for future work
would be to generalize the concepts in this paper to other problem domains where
transformations, like \csp\ to \sat, exist.

\paragraph{Acknowledgements.}
This work is supported by Science Foundation Ireland (SFI) Grant 10/IN.1/I3032 and FP7 FET-Open Grant 284715.
The Insight Centre for Data Analytics is supported by SFI
 Grant SFI/12/RC/2289.

\bibliographystyle{splncs03}
\bibliography{proteus}

\end{document}